\documentclass[letterpaper, 10 pt, conference]{ieeeconf}  

\IEEEoverridecommandlockouts                              

\usepackage{graphicx}  
\usepackage{amsmath} 
\usepackage{amssymb}  
\usepackage{booktabs}
\usepackage{array} 
\newcolumntype{w}[2]{>{\centering\arraybackslash}m{#2}}
\usepackage{xcolor}  

\makeatletter
\let\NAT@parse\undefined
\newlength{\figurecaptiongap}
\renewcommand{\@IEEEfigurecaptionsepspace}{\vskip\figurecaptiongap\relax}
\makeatother

\usepackage[colorlinks]{hyperref}
\usepackage{marvosym} 

\title{\LARGE \bf
ECHO: Event-Augmented Context with Hindsight and Outlook for Wrist-Only Manipulation
}

\author{\authorblockN{Xinyue Wang$^{1*}$, Yicheng Jiang$^{1*}$,  Zesen Gan$^{1}$, Junhao He$^{2}$, Jiaxu Wang$^{3}$,\\Junhao Li$^{1}$, Jingtao Zhang$^{1}$, Tianlun He$^{3}$, Jianan Wang$^{4}$, Isabel Guan$^{1,5\text{\Letter}}$ and Qiming Shao$^{1\text{\Letter}}$}%
\thanks{$^{*}$Contributed equally co-first authors, order determined by coin toss.}%
\thanks{$^{\text{\Letter}}$Corresponding authors:\newline\hspace*{2em}{\tt eeguan@ust.hk}, {\tt eeqshao@ust.hk}.}%
\thanks{$^{1}$The Hong Kong University of Science and Technology}%
\thanks{$^{2}$The University of British Columbia}%
\thanks{$^{3}$MMLab, The Chinese University of Hong Kong}%
\thanks{$^{4}$Astribot}%
\thanks{$^{5}$ZENBOT}%
}

\begin{document}


\maketitle
\thispagestyle{empty}
\pagestyle{empty}

\begin{abstract}

Learning-based manipulation policies relying on RGB cameras
often suffer from degraded observations under extreme
exposure. Event cameras mitigate this degradation by
asynchronously detecting pixel-level intensity changes to
offer a high dynamic range. However, their observations
heavily depend on camera placement, as fixed cameras miss
static scene content while wrist-mounted camera motion
causes previously visited regions to leave the field of
view. To address these spatial-temporal limitations, we
present ECHO (Event-augmented Context with Hindsight and
Outlook), a wrist-only latent world action model that
encodes wrist events into compact motion representations to
provide temporal and spatial context for policy reasoning.
Specifically, ECHO utilizes a pretrained event encoder to
explain visual-feature changes between frames. Its
hindsight module preserves the gripper trajectory with past
event stream as addressable off-camera context.
Concurrently, the outlook module introduces learnable event
foresight queries supervised to anticipate the event window
for future actions, enabling the policy to predict upcoming
scene changes. Evaluated on wrist-only RLBench tasks, ECHO
outperforms RGB and RGB+event baselines by 20.6 and 12.0
percentage points under normal lighting, and by 14.6 and
11.3 points under severe exposure drops, respectively,
while also surpassing RGB references using a third-person
camera. Real-world experiments with a wrist-mounted event
camera validate that ECHO outperforms RGB-only and
RGB+event baselines across multiple tasks under both
nominal and severely dark lighting. Project page is at https://echo-wam.github.io/.

\end{abstract}

\section{Introduction}

Recent years have seen remarkable progress in robotic manipulation, progressing from foundational imitation learning paradigms to sophisticated vision language action and world action models~\cite{rt2,openvla,cogact,pi0,pi05}. Although these models have achieved impressive success on diverse tasks, current evaluations mostly depend on standard RGB visual inputs captured in bright, ideal lighting conditions. Yet real-world deployments involving changing lighting conditions or rapid camera motion suffer from severe visual degradation, causing policies to fail with corrupted frames.

To address these visual failure modes, event cameras provide a compelling sensing modality by asynchronously report per-pixel brightness changes with microsecond timestamp resolution and high dynamic range~\cite{gallego2020}. They capture interactions such as approach, slip, and release when RGB frames are over-exposed, blur, or darken, provided the induced changes exceed the sensor's contrast threshold. Events therefore offer complementary motion cues for manipulation when RGB observations become unreliable.

However, event cameras respond only to brightness changes, introducing a trade-off between fixed and wrist-mounted configurations. When the camera is fixed, it misses static scene details that produce no brightness change. Mounting the camera on the wrist instead allows it to actively probe scene layout through motion, but the same motion also removes previously visited regions from view. This loss of visual context calls for information beyond the current observation. Existing event-augmented policies primarily incorporate event cues through visual fusion or residual action pathways~\cite{evla,eventvla}, without jointly organizing trajectory context, current event observations, and event foresight in the policy prefix.

We present ECHO, a wrist-only latent world action model that integrates events as a policy modality across past, present, and future. A pretrained encoder transforms pixel-level motion into latent representations that support both policy conditioning and future-event prediction. Looking back, a bounded trajectory memory combines visited gripper locations with current event features to provide \emph{off-camera context}; looking forward, \emph{event foresight} queries predict a dense representation of the event window spanned by the current action, anticipating scene changes before they are observed.Following the mainstream prefix-expert design, a vision-language backbone encodes observations and instructions into a token prefix, and a flow-matching expert generates continuous actions.

\begin{figure}[t]
  \centering
  \includegraphics[width=\columnwidth]{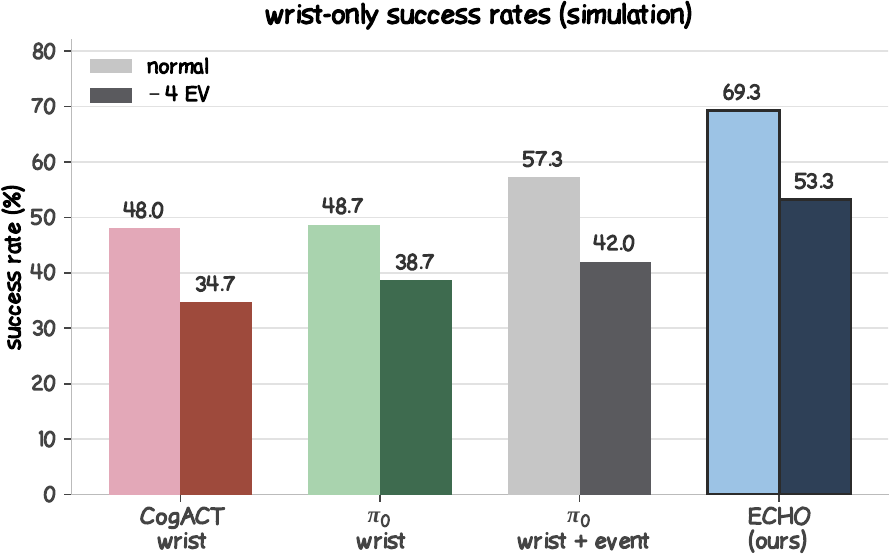}
  \caption{Success rates on the wrist-only RLBench under normal lighting and a $-4$ EV exposure setting. ECHO achieves the highest result in both conditions.}
  \label{fig:overall-results}
  \vspace{-15pt}
\end{figure}

Our contributions are summarized as follows:
\begin{itemize}
  \item A hybrid convolution-attention event encoder pretrained via cross-modal visual reconstruction and time-reversal contrastive learning to extract robust spatio-temporal motion priors.

  \item A wrist-only perception framework integrating \emph{trajectory memory} to keep out-of-view spatial cells addressable, and \emph{event foresight queries} to predict future action-conditioned event observations.

  \item Comprehensive evaluations on RLBench and real-robot platforms showing that ECHO consistently surpasses wrist RGB and RGB+event baselines under both standard and extreme low-light conditions (Fig.~\ref{fig:overall-results}).
\end{itemize}

\section{Related Work}

\subsection{Vision-language-action policies}
Vision-language-action (VLA) policies differ in action representation and decoding. RT-2 and OpenVLA generate action tokens from vision-language backbones~\cite{rt2,openvla}; CogACT couples a vision-language model to a diffusion action module~\cite{cogact}; and $\pi_0$ and $\pi_{0.5}$ generate continuous action chunks with flow-matching experts~\cite{pi0,pi05}. Following modern prefix-expert designs, ECHO augments an observation-and-language prefix with event-derived context to condition the flow-matching action expert.

\subsection{World models and latent action prediction}
World models predict future observations to support planning and policy learning. UniPi recovers actions from video plans through inverse dynamics~\cite{unipi}, while GR-1 and WorldVLA jointly predict future observations and actions~\cite{gr1,worldvla}. Ctrl-World supports policy evaluation and improvement through imagined rollouts~\cite{ctrlworld}. Other approaches focus on changes between observations, encoding them as compact latent actions~\cite{genie,lapo,lapa,clam,motus,lewm,villax} or continuous motion representations~\cite{como}. FLARE and VLA-JEPA further connect latent world modeling with policy learning through future-feature prediction~\cite{flare,vlajepa}. Following this latent world action modeling perspective, ECHO treats events as an efficient representation of observation changes. Its event encoder maps these signals into a compact latent space for policy reasoning, trajectory memory, and event foresight.

\subsection{Memory for partial observability and wrist views}
Memory supports manipulation when decisions depend on information no longer visible. MemoryVLA retrieves perceptual and semantic information from a memory bank~\cite{memoryvla}; HALO combines video question-answering supervision with sparse attention to retrieve task-relevant interaction history~\cite{halo}; and $\mu$VLA carries learnable memory tokens across timesteps through recurrent self-attention~\cite{muvla}.

Wrist-camera motion further motivates spatial memory. AtlasVLA combines a persistent voxel-hashed world state with a memory of robot state and task progress for manipulation from a single wrist camera~\cite{atlasvla}. Mem-World uses wrist-centered, surfel-indexed memory to retrieve historical views for consistent prediction in a multi-view world model~\cite{memworld}. ECHO quantizes the end-effector trajectory into visited spatial cells and embeds their positions together with the current event summary, keeping previously visited places accessible after they leave the frame.

\subsection{Event cameras in robot learning}
Event cameras provide high temporal resolution and dynamic range~\cite{gallego2020}, supporting obstacle avoidance, visual servoing, grasping, and visuomotor control~\cite{evdodgenet,sebvs,grasping_event,visuomotor_event}. Related work also explores early action prediction from event histories~\cite{earlypred}, event-language modeling~\cite{eventgpt}, and synchronized event and manipulation datasets~\cite{reassemble}. Recent event-augmented VLAs improve robustness to blur and low exposure by augmenting pretrained RGB policies with event cues. E-VLA fuses events into RGB inputs or visual features through overlays or hierarchical adapters~\cite{evla}, while Event-VLA introduces a gated residual fusion pathway after the VLA backbone\cite{eventvla}. ECHO instead treats events as a policy modality spanning past, present, and future within a shared prefix. Current event features tag previously visited gripper locations, observed event tokens support present perception, and foresight queries anticipate future event representations.

\begin{figure*}[t]
  \centering
  \includegraphics[width=\textwidth]{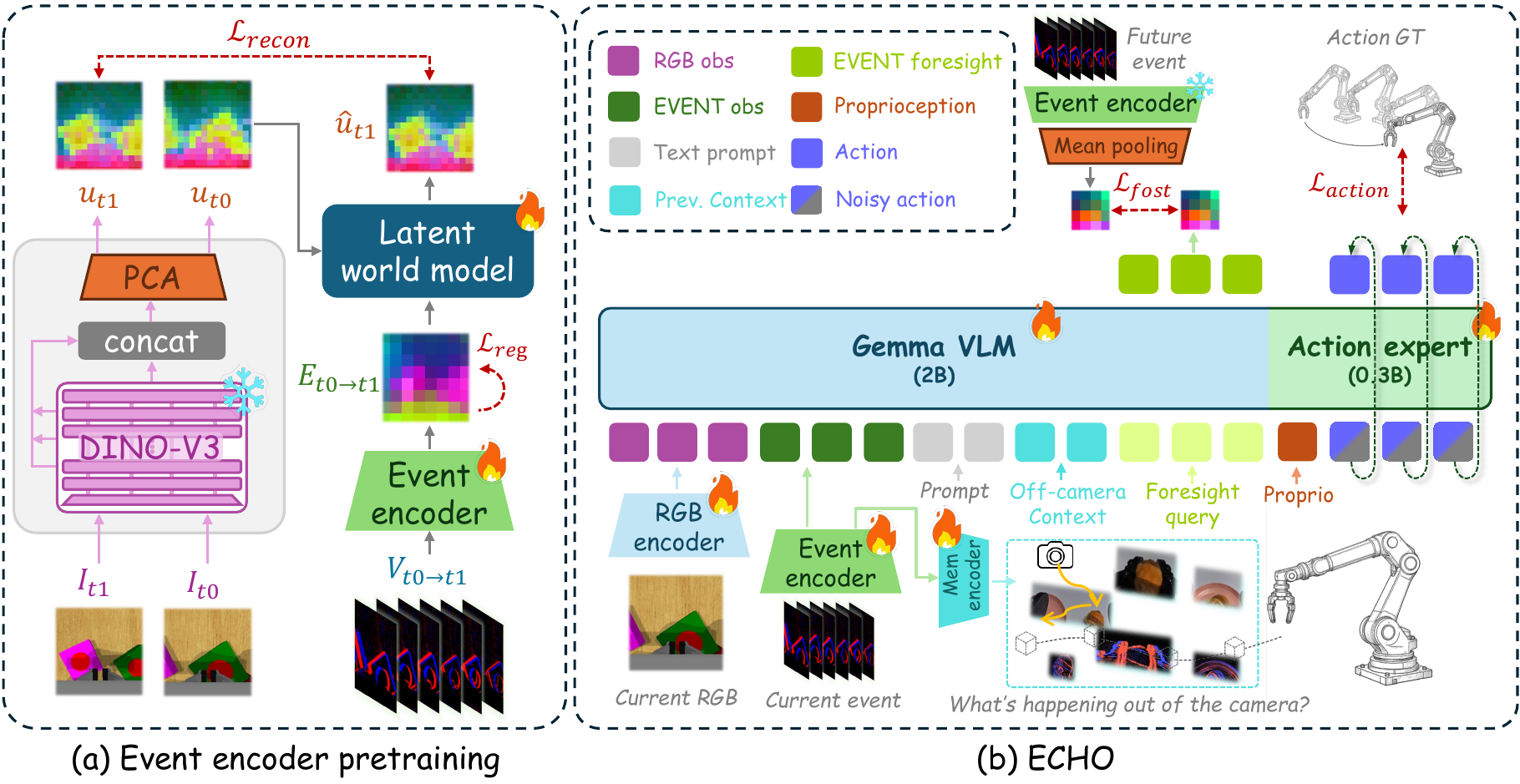}
  \caption{Overview of ECHO. (a) The event encoder is pretrained to represent the visual transition between consecutive RGB observations using frozen DINOv3 features and a latent world model. (b) During policy learning, wrist RGB, current event latent tokens $E_t$, language, off-camera context $C_t$ built from visited poses and current events, and foresight queries $F_t$ supervised by future event latents form the vision-language prefix. A flow-matching action expert then predicts continuous action chunks from this prefix and proprioception.}
  \vspace{-10pt}
  \label{fig:arch}
\end{figure*}

\section{Preliminaries}
\subsection{Policy interface and flow-matching action expert}
\label{sec:host}
Following common practice in modern manipulation policies, we adopt a prefix-expert architecture. At decision step $t$, let $I_t$ denote the wrist RGB observation, $\ell$ the language instruction, and $s_t$ the proprioceptive state. A vision-language backbone constructs the conditioning prefix as
\begin{equation}
 h_t = B_\psi(I_t,\ell).
\end{equation}
The action expert receives the proprioceptive state $s_t$ as a dedicated suffix token alongside the noisy action tokens.

The action expert conditions on this prefix and generates a chunk of $H$ actions. We write
\begin{equation}
 a_{t:t+H-1} \triangleq (a_t,\ldots,a_{t+H-1}) \in \mathbb{R}^{H\times d_a},
\end{equation}
where $d_a$ is the action dimension.

Flow matching trains the expert to transport Gaussian noise to the demonstrated action chunk~\cite{flowmatching,pi0}. For a training sample, draw $\varepsilon_a\sim\mathcal{N}(0,I)$ and interpolate with
\begin{equation}
 x_\tau = \tau\,\varepsilon_a+(1-\tau)\,a_{t:t+H-1},\qquad \tau\in[0,1],
\end{equation}
with target velocity
\begin{equation}
 v_a^\star=\varepsilon_a-a_{t:t+H-1}.
\end{equation}
At inference, numerical integration starts at $\tau=1$ and follows the learned velocity field toward $\tau=0$, yielding the action chunk.

\subsection{Event Stream and Voxel Representation}
An event $e_k=(x_k,y_k,t_k,p_k)$ records pixel coordinates, timestamp, and polarity $p_k\in\{-1,+1\}$. As illustrated in Fig.~\ref{fig:arch}, positive events (intensity increases) are shown in red, and negative events (intensity decreases) in blue. Events are triggered whenever the log intensity change crosses the contrast threshold~\cite{gallego2020}. 

To interface with a neural encoder, we accumulate asynchronous events over a decision window ending at $t$ into a signed voxel volume:
\begin{equation}
    V_t \in \mathbb{R}^{C \times H_e \times W_e}, \qquad C = 6.
\end{equation}
Specifically, events within the window (e.g., the previous completed keystep segment in simulation) are aggregated into $C=6$ equal-duration temporal bins. Signed event counts are halved and clipped to $[-1,1]$, preserving both polarity and coarse temporal ordering. For training in simulation, events are rendered via the ESIM simulator~\cite{esim}.

The event encoder $f_\theta$ maps this volume to $N$ event latent tokens,
\begin{equation}
 E_t=f_\theta(V_t)\in\mathbb{R}^{N\times D},\qquad N=64,
\end{equation}
which can be concatenated with the backbone prefix. ECHO trains $f_\theta$ to explain changes in frozen visual teacher features and uses the event latent for current perception, spatial context, and future-event supervision. The encoder architecture and pretraining are described in Sec.~\ref{sec:encoder}, and its prefix blocks in Secs.~\ref{sec:memory} and~\ref{sec:foresight}.

\section{Methodology}

\subsection{Overview}
ECHO extends the policy in Sec.~\ref{sec:host} with event features $E_t$, off-camera context $\mathcal{C}_t$, and foresight queries $\mathcal{F}_t$ (Fig.~\ref{fig:arch}). The resulting policy is
\begin{equation}
a_{t:t+H-1}=\pi\!\bigl(I_t,\,\ell,\,s_t,\,E_t,\,\mathcal{C}_t,\,\mathcal{F}_t\bigr).
\end{equation}
These prefix blocks provide present, past, and future context.
\begin{enumerate}
  \item \emph{Event features (present).} The encoder $f_\theta$ extracts a compact event latent $E_t$ from the observed volume $V_t$ (Sec.~\ref{sec:encoder}).
  \item \emph{Off-camera context (past).} The gripper trajectory is quantized into visited spatial cells, and $M$ context slots combine cell positions with the current event summary to keep visited places addressable outside the wrist view (Sec.~\ref{sec:memory}).
  \item \emph{Event foresight (future).} Learnable queries $\mathcal{F}_t$ are supervised to predict the latent representation of the future event window covered by the planned action (Sec.~\ref{sec:foresight}).
\end{enumerate}
The prefix contains image, event, language, context, and foresight tokens in order. The action expert receives a proprioceptive state token and $H$ noisy action tokens. Implementation settings are given in Sec.~\ref{sec:implementation}.

\begin{figure*}[t]
  \centering
  \includegraphics[width=\textwidth]{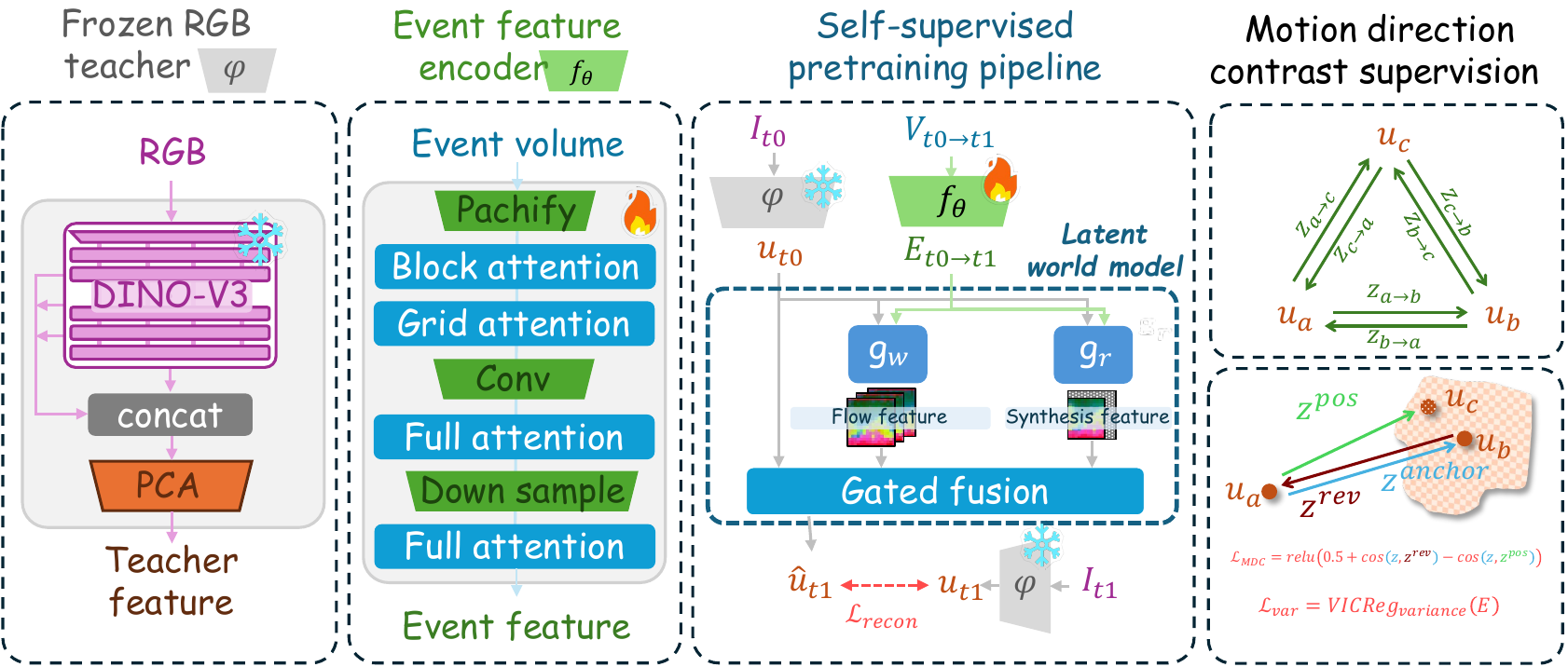}
  \caption{Event encoder pretraining. The event volume is encoded into an event latent that can (i) reconstruct the frozen RGB teacher's transition through a warp and synthesis head, and (ii) separate true motion window from its reversed twin under a motion-direction contrast (MDC).}
  \vspace{-15pt}
  \label{fig:encoder}
\end{figure*}

\subsection{Event feature encoder and pretraining}
\label{sec:encoder}
To capture fine-grained interaction dynamics, we design a convolution-attention encoder $f_\theta$ (Fig.~\ref{fig:encoder}) that processes a wrist event volume $V_t$ into a compact spatial latent grid:
\begin{equation}
E_t = f_\theta(V_t) \in \mathbb{R}^{N \times D},
\end{equation}
where $N$ denotes the number of spatial tokens and $D$ is their feature dimension.

We pretrain $f_\theta$ through a latent world model formulation, enabling event representations to predict visual-feature transitions and capturing temporal directionality under a frozen RGB teacher $\phi$ (DINOv3~\cite{dinov3}). Specifically, given $\mathbf{u}_t=\phi(I_t)$ and $\mathbf{u}_{t+\Delta}=\phi(I_{t+\Delta})$, the event latent explains their transition while $\mathbf{u}_t$ provides appearance. Multi-layer teacher features are aggregated, compressed, and spatially aligned to match the event token grid, grounding event tokens in rich visual semantics.
\emph{Visual change.} A warp head predicts per-token displacements and backward-warps the current teacher features as
\begin{equation}
W=\mathrm{Warp}\bigl(\mathbf{u}_t,\,g_{\mathrm{w}}(E_t)\bigr).
\end{equation}
A synthesis branch applies attention to $\mathbf{u}_t$, injecting event features at each block to produce
\begin{equation}
Q=g_{\mathrm{s}}\bigl(E_t,\,\mathbf{u}_t\bigr).
\end{equation}
A gate maps $[\mathbf{u}_t,W,Q]$ to two logits per token and applies a softmax across branches. With warp weight $\alpha$, the prediction is
\begin{equation}
\hat{\mathbf{u}}_{t+\Delta}=\alpha\odot W+(1-\alpha)\odot Q.
\end{equation}
Reconstruction targets the full future teacher feature using SmoothL1. Motion is emphasized through detached weights $w_i=d_i/\mathrm{mean}_j d_j$, where $d_i=\lVert(\mathbf{u}_{t+\Delta})_i-(\mathbf{u}_t)_i\rVert_2$. The reconstruction loss is
\begin{equation}
\mathcal{L}_{\mathrm{rec}}=\frac{1}{H_\phi W_\phi}\sum_{i=1}^{H_\phi W_\phi} w_i\,\mathrm{SmoothL1}\bigl((\hat{\mathbf{u}}_{t+\Delta})_i,\,(\mathbf{u}_{t+\Delta})_i\bigr),
\end{equation}
where $H_\phi$ and $W_\phi$ are the spatial dimensions of the teacher feature grid, so the loss averages over its spatial tokens.

The reversed branch swaps the frames and uses $V^{\mathrm{rev}}$ (defined below) with the same weights and loss to reconstruct $\mathbf{u}_t$, yielding $\mathcal{L}_{\mathrm{rec}}^{\mathrm{rev}}$.

\textit{Motion-direction contrast.} To encourage the encoder to capture motion directionality, we process an anchor event volume $V^{\mathrm{anc}}$, a duration-jittered positive view $V^{\mathrm{pos}}$, and a time-reversed view $V^{\mathrm{rev}} = -\mathrm{Flip}(V^{\mathrm{anc}})$ created by reversing temporal bins and inverting event polarity. For each view $v\in\{\mathrm{anc},\mathrm{pos},\mathrm{rev}\}$, the encoder produces $E^v=f_\theta(V^v)$, and each token is $\ell_2$-normalized as $z_i^v=E_i^v/\lVert E_i^v\rVert_2$ for the contrastive loss. Spatial selection of informative regions is achieved by max-pooling teacher feature change magnitudes $\lVert\mathbf{u}_{t+\Delta}-\mathbf{u}_t\rVert$ onto the event token grid, which identifies the set of most active tokens $\mathcal{M}$. The motion-direction contrastive (MDC) loss then enforces a margin $m$ between the positive and time-reversed views:
\begin{equation}
\mathcal{L}_{\mathrm{MDC}} = \frac{1}{|\mathcal{M}|} \sum_{i \in \mathcal{M}} \bigl[ \cos(z^{\mathrm{anc}}_i, z^{\mathrm{rev}}_i) - \cos(z^{\mathrm{anc}}_i, z^{\mathrm{pos}}_i) + m \bigr]_{+}.
\end{equation}

\emph{Variance Regularization.} We apply a batch-wise variance floor to the event encoder outputs $E$ before $\ell_2$ normalization:
\begin{equation}
\mathcal{L}_{\mathrm{var}}
=\frac{1}{ND}\sum_{i=1}^{N}\sum_{c=1}^{D}\bigl[\tau-\mathrm{std}_{b}(E_{b,i,c})\bigr]_{+},
\end{equation}
where $E_{b,i,c}$ denotes channel $c$ of token $i$ in batch sample $b$, and $\tau$ is the target standard deviation threshold.
The complete pretraining objective combines spatial reconstruction, temporal contrast, and feature variance regularization as
\begin{equation}
\mathcal{L}_{\mathrm{pre}}=\mathcal{L}_{\mathrm{rec}}+\mathcal{L}_{\mathrm{rec}}^{\mathrm{rev}}+\lambda_{\mathrm{var}}\,\mathcal{L}_{\mathrm{var}}+\lambda_{\mathrm{MDC}}\,\mathcal{L}_{\mathrm{MDC}},
\label{eq:pre}
\end{equation}
where $\lambda_{\mathrm{var}}$ and $\lambda_{\mathrm{MDC}}$ weight variance regularization and motion contrast. Both constraints act directly on the event latent tokens provided to the downstream policy.

\subsection{Off-camera context}
\label{sec:memory}

To retain spatial context after visited regions drift out of the wrist camera's field of view, we introduce a bounded trajectory memory. Given the end-effector position $p_\tau \in \mathbb{R}^3$ at step $\tau$ relative to a fixed origin $p_0$, we quantize the continuous motion into discrete spatial cells:
\begin{equation}
c_\tau = \mathrm{round}\!\left( \frac{p_\tau - p_0}{\delta} \right) \in \mathbb{Z}^3,
\end{equation}
where $\delta$ is the grid resolution and $q_c = p_0 + \delta c$ defines the cell center. The memory buffers the first $K$ unique cells in order of visitation. To construct the final hindsight context, we retrieve the prefix sequence of length $M$, beginning with the origin cell $c_0$.
Each context token combines a position embedding with the current event summary $\bar{e}_t=\frac{1}{N}\sum_{n=1}^{N}(E_t)_n\in\mathbb{R}^{D}$ as
\begin{equation}
m_c=\psi(q_c)+W_e\,\bar{e}_t,
\end{equation}
where $\psi$ is an MLP and $W_e$ projects to backbone width. Visited cells persist across steps, while the shared event term is recomputed from the current window.


The off-camera context block $\mathcal{C}_t=(m_{(1)},\ldots,m_{(M)})$ follows language tokens and precedes foresight queries. Image, event, and language tokens share an attention segment; $\mathcal{C}_t$, $\mathcal{F}_t$, and the action expert each begin a new one. Tokens attend bidirectionally within each segment and to all preceding segments, and the unused context slots are zero-padded.



\vspace{-5pt}

\subsection{Event Foresight Queries}
\label{sec:foresight}

To capture future action-conditioned dynamics, we append $N_f$ learnable tokens $\mathcal{F}_t = Q + P + \tau_{\mathrm{type}} \in \mathbb{R}^{N_f \times d_{\mathrm{vlm}}}$ to the observation prefix. These tokens attend over the entire observation prefix (including $I_t, \ell$, current event tokens, and $\mathcal{C}_t$).

During training, foresight queries regress the future event latent $\mathbf{f}_t$ generated by a frozen event encoder over the action horizon $H$ spanned by $a_{t:t+H-1}$. This encoder output serves as the target of an MSE loss with stop-gradient
\begin{equation}
\mathcal{L}_{\mathrm{foresight}}=\bigl\|\,\mathrm{Readout}(\mathcal{F}_t)-\mathrm{sg}(\mathbf{f}_t)\,\bigr\|_2^2 ,
\end{equation}
where $\mathbf{f}_t$ is obtained by spatial average pooling of the frozen encoder's event latent tokens into $N_f$ target tokens. Samples lacking valid future event windows are pad-masked and excluded from the loss.

\begin{equation}
\mathcal{L}=w(\mathrm{step})\,\mathcal{L}_{\mathrm{action}}+\lambda_f\,\mathcal{L}_{\mathrm{foresight}} .
\end{equation}

\begin{figure}[t]
  \centering
  \includegraphics[width=\columnwidth]{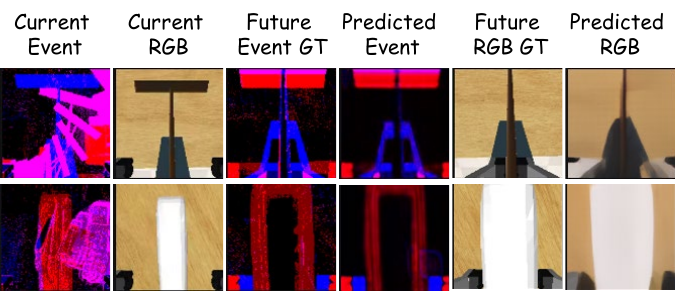}
  \vspace{-15pt}
  \caption{Qualitative visualization of future predictions. Current event and RGB observations are shown with future ground truth and decoded predictions. For visualization only, lightweight ViT heads decode events from foresight queries and reconstruct RGB images from future teacher features predicted by the latent world model conditioned on current teacher features and foresight queries.}
  \vspace{-15pt}
  \label{fig:predict}
\end{figure}

Fig.~\ref{fig:predict} visualizes event foresight through decoded future events and RGB reconstructions. The decoded events capture the main polarity patterns and motion locations, while the RGB reconstructions preserve the future object layout with smoother details than the ground truth.

\begin{figure}[t]
  \centering
  \includegraphics[width=\columnwidth]{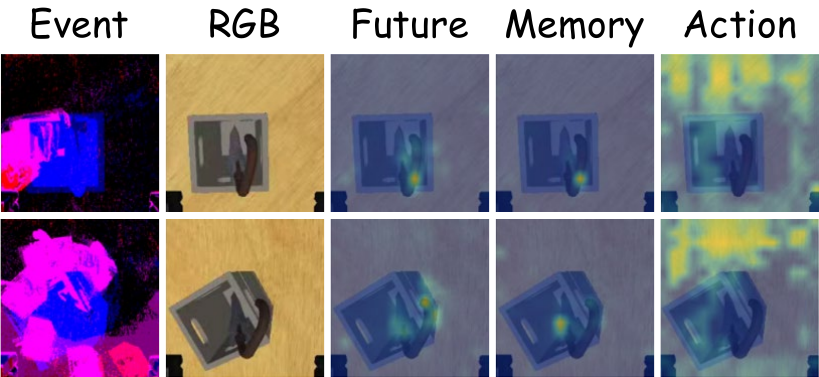}
  \vspace{-15pt}
  \caption{Qualitative attention visualization of different ECHO components. Under large wrist-camera motion where action tokens attend heavily to background movement, memory tokens concentrate precisely at gripper--object contact points. Meanwhile, foresight queries cover broader surrounding regions to anticipate interaction dynamics.}
  \vspace{-15pt}
  \label{fig:attention}
\end{figure}

Fig.~\ref{fig:attention} shows that trajectory memory tokens and foresight queries attend to interaction-relevant regions when action-expert attention is dominated by apparent background motion. The memory and foresight attention align with their intended functions, for spatial grounding and anticipating interaction dynamics.


\section{Experiments}
\label{sec:exp}

\subsection{Implementation details}
\label{sec:implementation}

We evaluate our approach on six representative RLBench tasks~\cite{rlbench} under a wrist-camera-only setting, utilizing wrist RGB images and the wrist event stream. For each task, we utilize episodes $0$ to $99$ for training, and episodes $100$ to $124$ for evaluation. 

The policy is trained for $20\text{k}$ steps with a global batch size of 32 across two H20 GPUs. The event encoder, foresight queries, their shared readout, and context projections are jointly trained with a learning rate of $1\times 10^{-4}$. The encoder starts from its pretrained weights (Sec.~\ref{sec:encoder}), while a frozen copy provides the foresight target. The queries are randomly initialized. Other modules are initialized from $\pi_0$.


 Following Event-VLA~\cite{eventvla}, during policy training, the wrist RGB input is randomly pad-masked with a probability of $0.3$ while retaining the complete event stream for all the methods incorporating event inputs. At closed-loop evaluation, event streams between consecutive RGB frames are generated online via the same dense ESIM pipeline used for training data generation~\cite{esim}. Simulation uses ESIM events, and the real-robot experiments in Sec.~\ref{sec:real} use a physical DVSync camera from DVSense~\cite{dvsense}.

\textbf{Model structure.} ECHO pairs a vision-language prefix backbone with a flow-matching action expert; both blocks are initialized from $\pi_0$~\cite{pi0}. The event encoder has approximately 60M parameters and maps six-channel event volumes to $N{=}64$ latent tokens of width $D{=}1024$, arranged on an $8{\times}8$ grid in row-major order.

The context module utilizes a spatial resolution of $\delta{=}0.025\,\mathrm{m}$, retains up to $K{=}64$ visited cells, and reads $M{=}16$ context slots in first-seen order. Cell positions are embedded with a two-layer MLP. The foresight block appends $N_f{=}16$ queries, supervised by future event tokens pooled from an $8{\times}8$ to a $4{\times}4$ grid using $2{\times}2$ average pooling. We set the action horizon to $H{=}1$ for RLBench keystep prediction following ~\cite{hybridvla} and $H{=}16$ for real-robot deployment (Sec.~\ref{sec:real}).

\textbf{Encoder pretraining.} DINOv3-ViT-L/16 processes images with a resolution of $256{\times}256$ pixels. Features from layers $\{6,12,18,24\}$ are concatenated, projected by offline PCA to 256 dimensions, LayerNormalized per token, and averaged over four one-pixel shifts to reduce patch-grid aliasing. Event tokens are aligned to the resulting $16{\times}16$ teacher grid by nearest replication, each covering a $2{\times}2$ block. Training pairs span $\Delta\in\{1,2,3\}$ keystep segments. We use SmoothL1 with $\beta{=}0.1$, duration jitter up to $20\%$, the top-$10\%$ active tokens for MDC, a variance threshold $\tau{=}0.5$, and $\lambda_{\mathrm{var}}{=}1$. Both reconstruction branches have unit weight.

\subsection{Main results}

\begin{table*}[t]
\centering
\caption{Success rate (\%) on six RLBench tasks using the wrist-only camera and event protocol of Sec.~\ref{sec:exp}. Results are reported under normal lighting (left) and a $-4$ EV exposure shift (right).}

\vspace{-10pt}

\label{tab:main}
\scriptsize
\setlength{\tabcolsep}{2pt}

\begin{tabular}{l|c@{\hspace{1.5pt}}c@{\hspace{1.5pt}}c@{\hspace{1.5pt}}c@{\hspace{1.5pt}}c|*{6}{@{\hspace{1.5pt}}w{c}{1.8em}@{\textcolor{black!40}{$|$}}w{c}{1.8em}}|@{\hspace{1.5pt}}r@{\textcolor{black!40}{$|$}}l}
\toprule
Method & Event & Pretrained & Traj-ctx & Event-ctx & Foresight & \multicolumn{2}{c}{laptop} & \multicolumn{2}{c}{toilet} & \multicolumn{2}{c}{fridge} & \multicolumn{2}{c}{umbrella} & \multicolumn{2}{c}{frame} & \multicolumn{2}{c}{plants} & \multicolumn{2}{c}{SR (\%)} \\
& & & & & & \textcolor{black!60}{N} & \textcolor{black!60}{D} & \textcolor{black!60}{N} & \textcolor{black!60}{D} & \textcolor{black!60}{N} & \textcolor{black!60}{D} & \textcolor{black!60}{N} & \textcolor{black!60}{D} & \textcolor{black!60}{N} & \textcolor{black!60}{D} & \textcolor{black!60}{N} & \textcolor{black!60}{D} & \textcolor{black!60}{N} & \textcolor{black!60}{D} \\
\midrule
$\pi_0$ (Front only) & & & & & & 16 & 17 & \underline{24} & 17 & \textbf{25} & 18 & 12 & 3 & 5 & \underline{4} & 1 & \underline{6} & 55.3\,\textcolor{red}{($-$2.0)} & 43.3\,\textcolor{red}{(+1.3)} \\
$\pi_0$ (Front + wrist) & & & & & & 19 & 9 & \underline{24} & 17 & 20 & 20 & \underline{15} & 5 & \textbf{11} & 3 & \underline{8} & 2 & 64.7\,\textcolor{red}{(+7.4)} & 37.3\,\textcolor{red}{($-$4.7)} \\
$\pi_0$ (Wrist only) & & & & & & 18 & 17 & \underline{24} & 12 & 20 & 20 & 4 & 5 & 3 & 1 & 4 & 3 & 48.7\,\textcolor{red}{($-$8.6)} & 38.7\,\textcolor{red}{($-$3.3)} \\
CogACT (Wrist only) & & & & & & \underline{23} & 10 & 20 & 17 & 18 & 19 & 0 & 0 & 8 & \underline{4} & 3 & 2 & 48.0\,\textcolor{red}{($-$9.3)} & 34.7\,\textcolor{red}{($-$7.3)} \\
$\pi_0$ + event & \checkmark & \checkmark & & & & \textbf{24} & \textbf{20} & \underline{24} & 14 & 22 & 17 & 5 & \underline{9} & 3 & 0 & \underline{8} & 3 & 57.3\,\textcolor{red}{(+0.0)} & 42.0\,\textcolor{red}{(+0.0)} \\
\midrule
ECHO (action-expert event foresight) & \checkmark & \checkmark & \checkmark & \checkmark & \checkmark$^{\ast}$ & 8 & 10 & 23 & 17 & 21 & \textbf{22} & 5 & 3 & 6 & 2 & 7 & 1 & 46.0\,\textcolor{red}{($-$11.3)} & 36.7\,\textcolor{red}{(-5.3)} \\
\midrule
ECHO w/o pretrained encoder, foresight & \checkmark & & \checkmark & \checkmark & & 19 & 18 & 23 & \underline{20} & 20 & \underline{21} & 8 & 1 & 8 & 0 & 6 & 0 & 56.0\,\textcolor{red}{($-$1.3)} & 40.0\,\textcolor{red}{($-$2.0)} \\
ECHO w/o foresight & \checkmark & \checkmark & \checkmark & \checkmark & & 22 & 18 & \underline{24} & 16 & \underline{24} & \underline{21} & 10 & 4 & 6 & 2 & \textbf{12} & 3 & \underline{65.3}\,\textcolor{red}{(+8.0)} & 42.7\,\textcolor{red}{(+0.7)} \\
ECHO w/o context & \checkmark & \checkmark & & & \checkmark & 22 & 18 & \textbf{25} & 14 & 21 & \textbf{22} & 5 & 5 & 7 & \textbf{5} & 1 & 2 & 54.0\,\textcolor{red}{($-$3.3)} & 44.0\,\textcolor{red}{(+2.0)} \\
ECHO w/o event context & \checkmark & \checkmark & \checkmark & & \checkmark & 21 & \textbf{20} & 23 & \underline{22} & \underline{24} & \textbf{22} & 13 & 6 & 8 & 3 & 3 & 1  & 60.7\,\textcolor{red}{(+3.4)} & \underline{48.7}\,\textcolor{red}{(+6.7)} \\
\textbf{ECHO (full)} & \checkmark & \checkmark & \checkmark & \checkmark & \checkmark & \textbf{24} & \textbf{20} & \textbf{25} & \textbf{23} & 20 & \underline{21} & \textbf{17} & \textbf{11} & \underline{10} & \underline{4} & \underline{8} & 1 & \textbf{69.3}\,\textcolor{red}{(+12.0)} & \textbf{53.3}\,\textcolor{red}{(+11.3)} \\
\bottomrule
\end{tabular}
\vspace{-10pt}
\end{table*}

In Table~\ref{tab:main}, checkmarks identify the active components, with Traj-ctx and Event-ctx denoting trajectory and event context tokens. The $^{\ast}$ variant places foresight queries in the action expert rather than the prefix. The highest value in each column and lighting condition is bold, and the next highest is underlined. Red parenthesized SR deltas are measured relative to the $\pi_0$ + event baseline.

\textbf{Overview.} On the six-task wrist-only benchmark (Table~\ref{tab:main}), ECHO achieves a \textbf{69.3\%} average success rate under normal lighting, ranking first among wrist-only methods. It outperforms the RGB-only wrist baselines $\pi_0$ (48.7\%) and CogACT (48.0\%), the event-augmented wrist baseline $\pi_0$ + event (57.3\%), and the third-person RGB references (55.3\% front-only, 64.7\% front+wrist). Under an extreme $-4$~EV exposure shift (about $1/16$ illuminance, right side of Table~\ref{tab:main}), ECHO maintains \textbf{53.3\%} success, ahead of $\pi_0$ + event (42.0\%), the wrist-only RGB baselines (38.7\% $\pi_0$, 34.7\% CogACT), and the third-person RGB references (43.3\% front-only, 37.3\% front+wrist).
\subsection{Ablation Study}
\label{sec:exp-ablation}
\textbf{Results and Analysis.} Table~\ref{tab:main} reports ablations of the ECHO components.

\begin{figure}
    \centering
    \includegraphics[width=\linewidth]{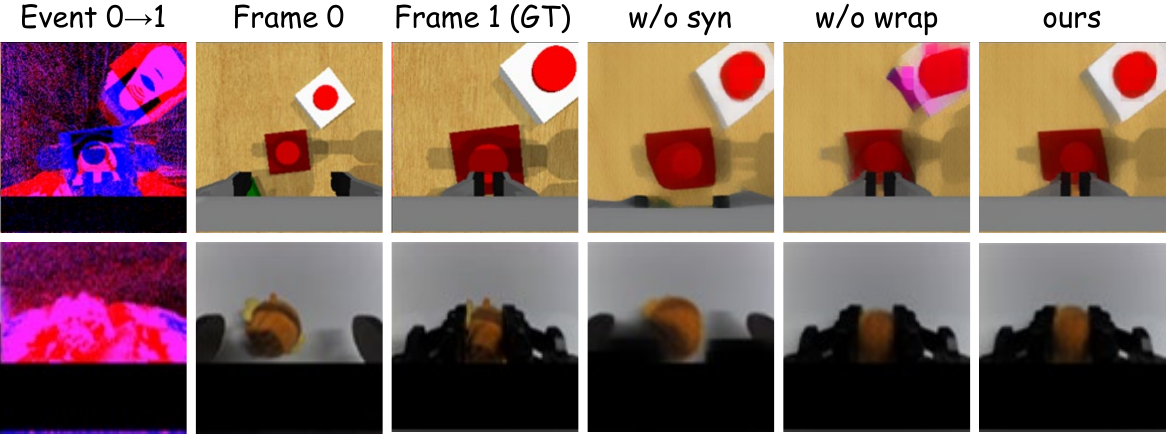}
    \caption{Qualitative reconstruction examples of the pretrained event encoder, on a simulation sequence and a real-robot sequence.}
    \vspace{-15pt}
    \label{fig:future-visualization}
\end{figure}

\textit{1) Pretrained Event Representation.} We first compare the wrist-only RGB baseline with $\pi_0$ + event, which adds pretrained event tokens without context or foresight. This increases average success from 48.7\% to 57.3\% under normal lighting and from 38.7\% to 42.0\% under severe darkness ($-4$~EV), indicating that current event observations provide useful cues for action generation (Table~\ref{tab:main}). To isolate the contribution of pretraining, we then compare \emph{ECHO w/o foresight} with \emph{ECHO w/o pretrained encoder, foresight}. Both variants retain the trajectory and event context blocks and omit foresight queries, while the latter trains the event encoder from scratch during policy learning. Pretraining raises success from 56.0\% to 65.3\% under normal lighting and from 40.0\% to 42.7\% under darkness, giving gains of 9.3 and 2.7 percentage points, respectively. Fig.~\ref{fig:future-visualization} further separates the roles of the two reconstruction branches. The warp head transports current semantic features along event-induced displacements and captures the motion direction of content present in the current frame. The synthesis head can introduce newly visible information, although its output is less constrained to preserve the current semantic structure. In the real-robot example, variants containing the synthesis branch recover the changing gripper configuration, while the warp-only reconstruction does not. Together, these results indicate that latent-wm style encoder pretraining provides a useful initialization for downstream policy learning, with benefits even when RGB observations are well exposed.

\textit{2) Off-Camera Trajectory Memory.} Incorporating the trajectory memory block onto the event-input baseline (\emph{ECHO w/o foresight} vs.\ $\pi_0$ + event) boosts normal-light success from 57.3\% to 65.3\% (Table~\ref{tab:main}). The largest per-task gains appear in multi-stage tasks where the wrist camera temporarily loses sight of targets during workspace exit and re-entry, e.g., \emph{umbrella} ($5 \to 10$). However, under extreme darkness (Table~\ref{tab:main}), maintaining trajectory memory alone fails to prevent low-light failure, yielding limited gain over wrist RGB ($42.7\%$ vs. $38.7\%$).

\textit{3) Event Foresight Queries.} Event foresight queries serve as the primary mechanism for mitigating severe low-light degradation, boosting dark success from 42.7\% to 53.3\% ($+10.6\%$, Table~\ref{tab:main}), with substantial recoveries in contact-heavy tasks such as \emph{umbrella} ($4 \to 11$) and \emph{toilet} ($16 \to 23$). Notably, foresight queries depend on spatial grounding from the trajectory memory, training without context (\emph{ECHO w/o context}) drops normal-lighting success to 54.0\% (below the 57.3\% baseline). Furthermore, relocating foresight predictions from the prefix into the action expert (\emph{ECHO w/o prefix foresight}) lowers performance significantly, suggesting that event foresight functions more effectively as a prefix context than inside the action expert.

In summary, \textbf{pretrained event representations} provide fundamental motion priors, \textbf{trajectory memory} anchors persistent spatial representations when workspace regions leave the wrist field of view, and \textbf{event foresight} yields predictive cues that sustain policy execution under severe photometric degradation.

\subsection{Real-robot deployment}
\label{sec:real}

\begin{table}[t]
\centering
\caption{Real-robot success rate (\%) across tasks under normal and severe dark lighting ($20$ trials per cell). Red parenthesized deltas are relative to the $\pi_0$ + event baseline.}
\vspace{-10pt}
\label{tab:real_robot}
\footnotesize
\setlength{\tabcolsep}{1.8pt}
\begin{tabular}{llcccc}
\toprule
& & \multicolumn{2}{c}{Normal Lighting} & \multicolumn{2}{c}{Severe Dark Lighting} \\
\cmidrule(lr){3-4} \cmidrule(lr){5-6}
Method & Perception & pick-place & bell-ring & pick-place & bell-ring \\
\midrule
$\pi_0$ & wrist RGB & 80 \textcolor{red}{(+5)} & 15 \textcolor{red}{($-$15)} & 55 \textcolor{red}{($-$5)} & 5 \textcolor{red}{($-$10)} \\
$\pi_0$ + event & wrist RGB + event & 75 \textcolor{red}{(+0)} & 30 \textcolor{red}{(+0)} & 60 \textcolor{red}{(+0)} & 15 \textcolor{red}{(+0)} \\
\textbf{ECHO} & wrist RGB + event & \textbf{90} \textcolor{red}{(+15)} & \textbf{45} \textcolor{red}{(+15)} & \textbf{80} \textcolor{red}{(+20)} & \textbf{30} \textcolor{red}{(+15)} \\
\bottomrule
\end{tabular}
\end{table}

\begin{figure}[t]
  \centering
  \setlength{\figurecaptiongap}{1pt}
  \includegraphics[width=\columnwidth]{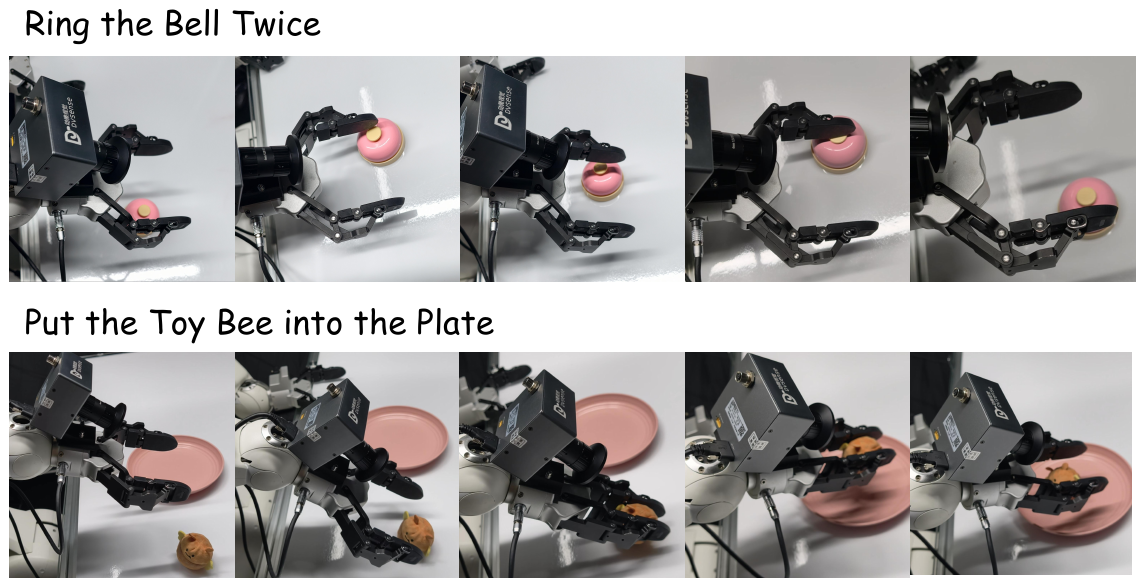}
  \caption{Real-robot setup for the tasks \emph{Ring the Bell Twice} and \emph{Put the Toy Bee into the plate}.}
  \label{fig:3rd}
  \vspace{-18pt}
\end{figure}

\textbf{Experimental setups.} The real-world setup uses an Astribot~S1~\cite{astribot_s1} (Fig.~\ref{fig:3rd}). The policy is deployed on an RTX~3090. Visual and event observations are collected from a wrist-mounted DVSync Event camera~\cite{dvsense}. We evaluate two tasks, \emph{Put the Toy Bee into the Plate} and \emph{Ring the Bell twice}. Each task provides $100$ demonstrations, $75$ recorded under normal lighting and $25$ under severe dark lighting.


 Each policy is run for $20$ trials per task, under normal lighting at $286$\,lux and dark lighting below $10$\,lux, set by a dimmable lamp above the workspace.

 In closed loop the policy is queried every $H{=}16$ RGB intervals and executes the returned chunk. The foresight queries and the context block are updated at this frequency. The foresight window is the event stream of the $H$ intervals the chunk spans, and the context block is updated once per step from the end-effector history since the episode start. The observed event latent is recomputed per RGB interval, encoding the last 16 segments, each represented as a six-bin volume (Sec.~\ref{sec:encoder}).



\begin{figure}[t]
  \centering
  \setlength{\figurecaptiongap}{1pt}
  \includegraphics[width=\columnwidth]{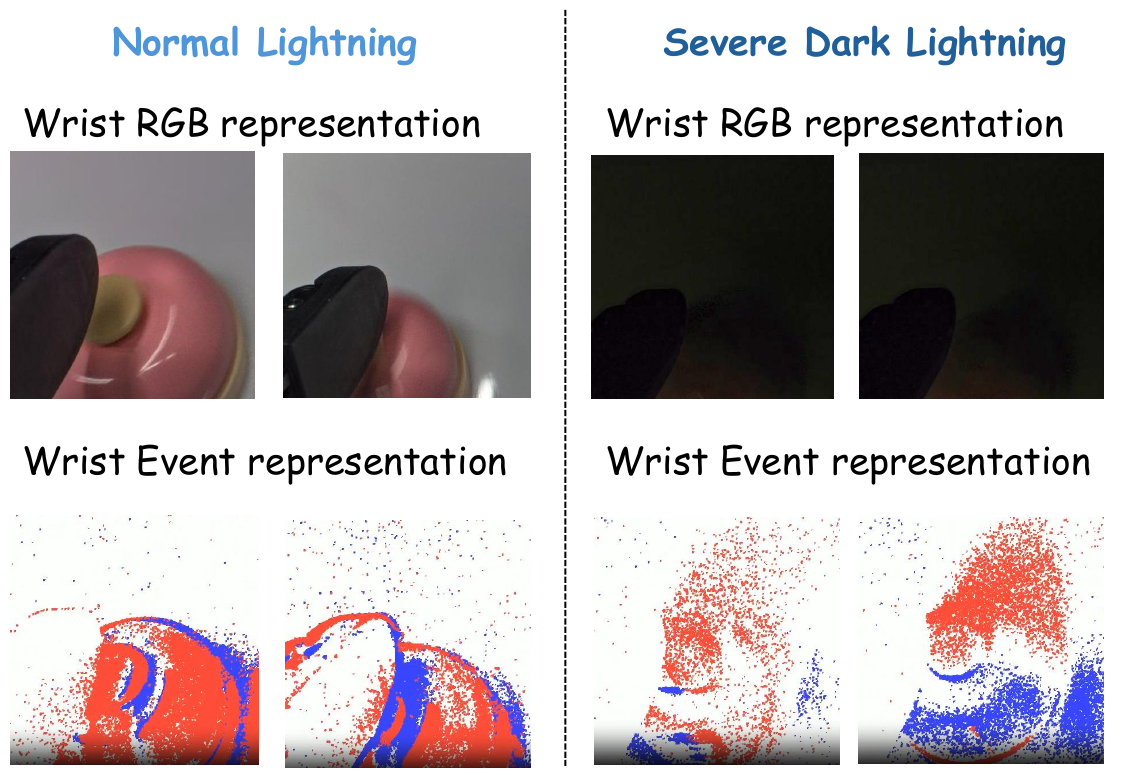}
  \caption{Real-robot execution of \emph{ring the bell twice} under normal and dark lighting. Wrist RGB (top) and events from the same interval (bottom) show contact between the gripper and bell.}
  \label{fig:real}
  \vspace{-18pt}
\end{figure}

\textbf{Results and Analysis.} 

\textit{1) Normal Lighting.} Across both real-robot tasks, ECHO consistently outperforms the wrist-only RGB baseline (Table~\ref{tab:real_robot}). The performance gap is most pronounced in \emph{bell-ring} ($45\%$ vs. $15\%$), where executing repeated strikes requires keeping track of the bell after the wrist-mounted camera moves past it. The spatial trajectory memory in ECHO keeps the target location addressable even when it leaves the camera frame.

\textit{2) Severe Dark Lighting.} As shown in Fig.~\ref{fig:real}, extreme low light severely corrupts RGB inputs, whereas the synchronous event stream robustly captures temporal brightness changes during physical interaction. Quantitatively, ECHO maintains strong reliability: on \emph{pick-place}, its performance margin over the baseline widens from $+10\%$ to $+25\%$ ($80\%$ vs. $55\%$, Table~\ref{tab:real_robot}). Notably, simply appending instant event features ($\pi_0 + \text{event}$) yields inconsistent results across tasks, confirming that raw event streams require explicit trajectory context and predictive grounding to handle lighting and view shifts effectively.

\section{Conclusions}

In this work, we introduce ECHO, an event-based framework for robot manipulation under challenging environments. ECHO leverages a pretrained event feature encoder to embed wrist-mounted event volumes into representation vectors that explicitly capture event-motion dynamics in a latent-wm style. Furthermore, its hindsight block injects gripper trajectories with event as off-camera context, while the outlook block optimizes foresight queries to predict future event horizons for action generation. Extensive evaluations on RLBench and two real-robot tasks demonstrate that ECHO consistently outperforms RGB-only baselines across both standard and severely degraded lighting conditions.
\section*{Acknowledgment}

Codex~\cite{codex} with ChatGPT (GPT-5.6~\cite{gpt56}) was used to assist with coding in this project, and the robot-arm illustration in Fig.~\ref{fig:arch} was generated with GPT Image 2~\cite{gptimage2}. All other content was created by the authors without generative models.

\bibliographystyle{IEEEtran}
\bibliography{refs}

\end{document}